%% file: root.tex
\documentclass[letterpaper, 10 pt, conference]{ieeeconf}  

\input{preamble}

\title{\LARGE \bf
Reinforcement Learning with Time-dependent Goals for Robotic Musicians
}

\author{Thilo Fryen$^{1*}$, Manfred Eppe$^{1*}$, Phuong D.H. Nguyen$^{1}$, Timo Gerkmann$^{2}$, Stefan Wermter$^{1}$
\thanks{$^*$equal contribution. $^{1}$Thilo Fryen, Manfred Eppe, Phuong D.H. Nguyen, and Stefan Wermter are with the Knowledge Technology group, Department  of  Informatics, University of Hamburg, Hamburg, Germany. $^{2}$Timo Gerkmann is with the Signal Processing group, Department  of  Informatics, University of Hamburg, Hamburg, Germany.
        {\tt\small \{6fryen, eppe, pnguyen, gerkmann, wermter\}@informatik.uni-hamburg.de}}%
}

\usepackage{xcolor}

\begin{document}

\maketitle
\thispagestyle{firstpage}
\pagestyle{empty}

\begin{abstract}
Reinforcement learning is a promising method to accomplish robotic control tasks. The task of playing musical instruments is, however, largely unexplored because it involves the challenge of achieving sequential goals -- melodies -- that have a temporal dimension. In this paper, we address robotic musicianship by introducing a temporal extension to goal-conditioned reinforcement learning: Time-dependent goals.
We demonstrate that these can be used to train a robotic musician to play the theremin
instrument.
We train the robotic agent in simulation and transfer the acquired policy to a real-world robotic thereminist. \newline Supplemental video: {\url{https://youtu.be/jvC9mPzdQN4}}
\end{abstract}


\section{Introduction}
\begin{figure}
    \centering
    \includegraphics[trim=30pt 10pt 0pt 0pt), clip,width=.95\columnwidth]{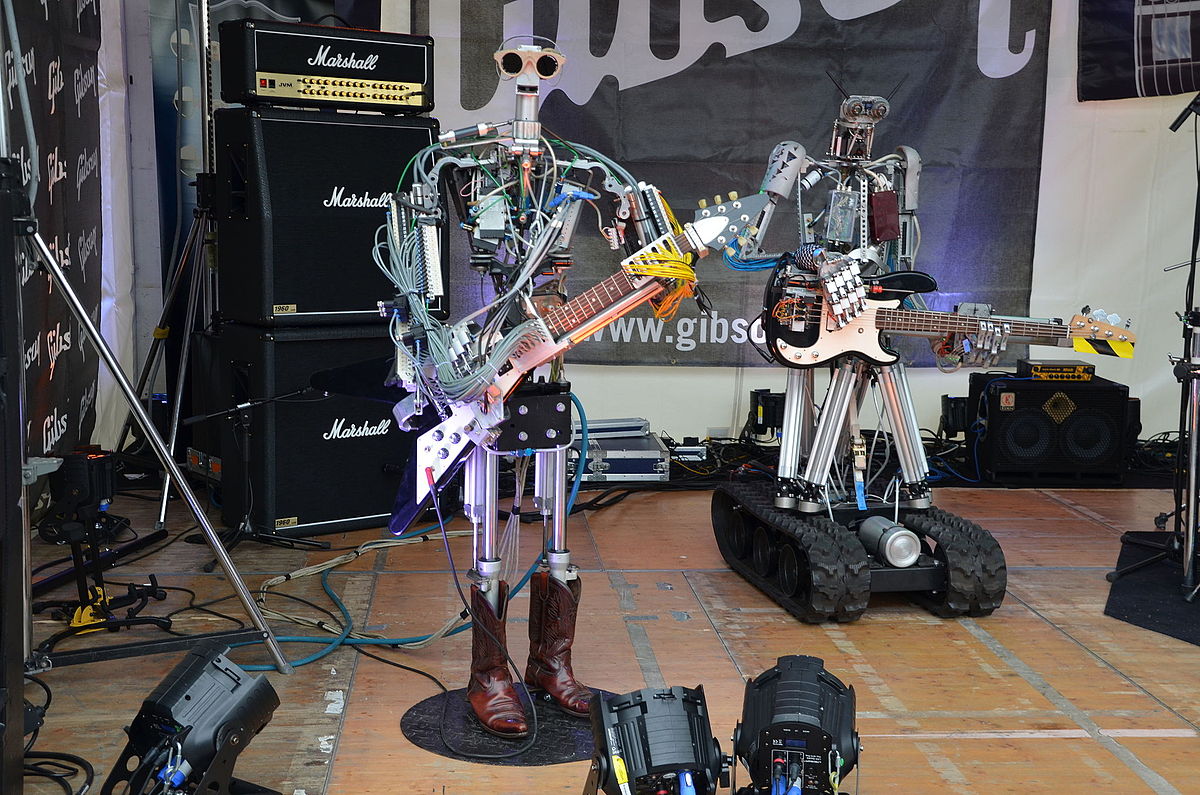}
    \hspace{-\columnwidth}\begin{minipage}[c]{.95\columnwidth}\footnotesize \color{white}{\hspace{.05\columnwidth}Photo by Torsten Maue, licensed under the terms of the cc-by-sa-2.0.}\end{minipage}
    \caption{The guitarist and bassist of the robotic metal band \emph{Compressorhead}.}
    \label{fig:compressorhead}
\end{figure}

Most contemporary musical performance robots rely on hard-coded motion trajectories to play musical instruments. 
For example, the robotic metal band \emph{Compressorhead} consists of five humanoid robots that play a guitar, a bass guitar, and drums using sophisticated pneumatic actuators (see \autoref{fig:compressorhead}). A problem with this approach is that engineers and artists need to calibrate and program the specific motion trajectories required to play the instruments manually. 
How can we realize a computational architecture that allows robotic musicians to learn to play their instruments so that artists and engineers do not need to program them manually?
A currently very prominent learning-based approach for robotic control that can potentially overcome this issue is reinforcement learning (RL). 
However, existing approaches have not been able to realize RL-driven robotic musicians because music is time-dependent: To play a melody, a robotic musician must hit the desired pitch of each note in a temporal sequence. A problem is that current reinforcement learning methods consider static goals or static reward functions that are independent of the temporal dimension of an episode. This renders them inappropriate for learning the temporal dynamics involved in controlling musical instruments. 

As our novel scientific contribution, we address this problem by extending static goal-conditioned reinforcement learning methods with \emph{time-dependent goals} (TDGs). Herein, we extend the static goal representation provided to the agent with a dynamically changing goal representation. We hypothesize that time-dependent goals enable robots to play musical instruments. 
We verify our hypothesis for the case of playing the theremin (Fig. \ref{fig:theremin}) and we provide a proof of concept by applying the method to a robotic theremin player. We also provide an empirical evaluation to show that the method is robust to acoustic noise and we compare different variations of the auditory preprocessing and control mechanisms. 

The remainder of this paper is constructed as follows. We provide a brief background and related work for musical robots and goal-conditioned reinforcement learning in section \ref{sec:relw}.
In section \ref{sec:meth}, we introduce the time-dependent goal and describe our theremin tone simulation. Then we describe our experiments in both the simulated and the real-world environment in section \ref{sec:expe}. We illustrate our architecture and the reinforcement learning components.
Our results section \ref{sec:resu} is divided into the comparative evaluation of different time- to frequency-domain transforms, action spaces, and robustness to different noise intensities. We also investigate hindsight experience replay to alleviate reward-sparsity. Furthermore, we provide a real-world proof-of-concept and transfer the approach to a real-world robotic theremin-playing robot. We conclude in section \ref{sec:conc}.

\section{Background and Related Work}
\label{sec:relw}

We combine robotic control through reinforcement learning with musical robots. 
There exist works that compose music with reinforcement learning methods \cite{Collins2008, LeGroux2010}, but to the best of our knowledge, there exists no prior work that uses reinforcement learning to control a physical robotic musician.

\subsection{Musical Robots}
The design of musical robots and the development of the corresponding algorithms constitutes an interesting challenge because of the many aspects that music incorporates.
Existing robotic musicians include the flute and saxophonist robots from Waseda University \cite{Solis}, the marimba-playing robot \textit{Shimon} from the Georgia Institute of Technology \cite{Shimon}, and the violinist robot from Ryukoku University \cite{Ryukoku_violin}. 
These and other prominent projects, like \textit{Compressorhead} \cite{compressorhead} and \textit{Automatica} \cite{automatica}, concentrate on the artistic and the engineering aspects of robotic musicianship. This leads to impressive performances and a relatively precise control of non-trivial instruments. However, large parts of the behavior of these robots are pre-programmed and depend on hand-engineered domain knowledge. 
A learning-based approach can simulate the individuality of human musicians better because they are not explicitly programmed but rather learn their individual way of playing the instrument. A learning-based approach also eliminates the need for background knowledge and the manual calibration of control mechanisms. 
However, the temporal dimension that inheres musical melodies and harmonies makes it difficult to apply existing learning-based methods, especially for instruments that are difficult to play. 
So how can we realize a learning-based approach that is suitable for learning a sequence of different pitches while also considering rhythm and timing? How can we avoid the manual calibration of robotic musicians?

\begin{figure}[ht]
         \centering
          \includegraphics[trim=0 0 0 -0.8cm, width=\linewidth]{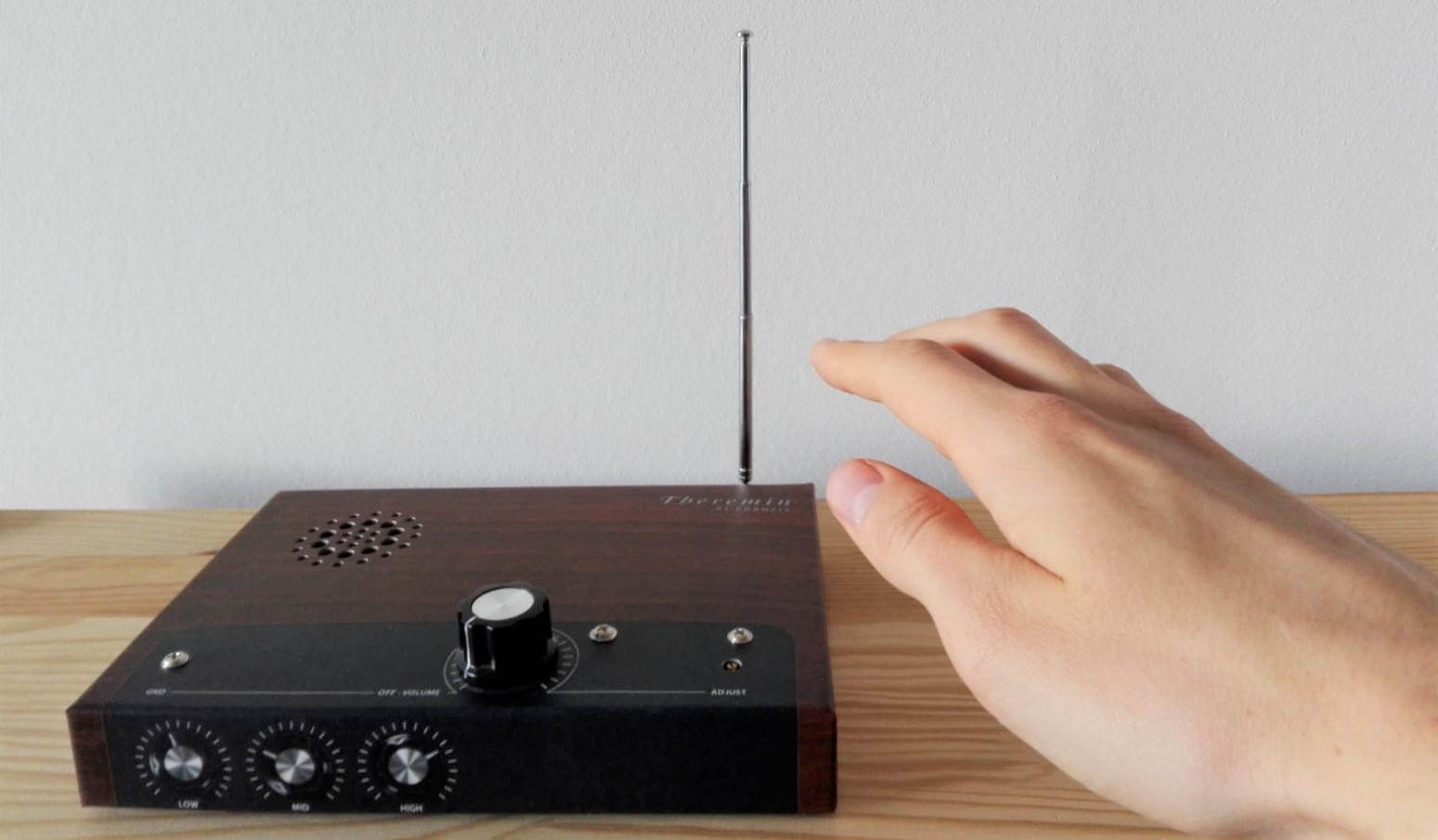}
         \caption{The physical theremin we used for our experiments. The pitch is controlled by adjusting one's distance to the theremin's antenna.}
          \label{fig:theremin}
\end{figure}

To address these challenging questions, we propose here a proof of concept with a theremin-playing robot. 
The theremin (see Fig. \ref{fig:theremin}) is an analog electronic instrument that emits a sine-like tone, where the pitch can be controlled by changing the distance of the musician's hand to the theremin's antenna. 
The simplicity of this design allows us to develop and to evaluate a temporally extended actor-critic reinforcement learning method in conjunction with domain randomization \cite{openai2019solving}. This enables the robot to learn to play temporally extended melodies and to quickly adapt to the actual properties of the environment without requiring a calibration phase. 
Previous robotic thereminists \cite{Mizumoto2009, Wu2010} use feed-forward control and therefore need a calibration phase to determine the distance-to-frequency function before the robot can play the theremin. 


\subsection{Actor-Critic and Goal-conditioned Reinforcement Learning}
In reinforcement learning, an agent interacts with an environment and learns to solve a problem. At each time step, the agent receives a representation of the environment's state \textit{s} and carries out an action \textit{a} according to its policy $\pi (s) = a$. 
The environment defines a reward function $r(s)$ that informs the agent about the success of its policy.
To be able to learn the optimal policy and maximize the rewards, the agent needs to know the best action for each state.
It approximates an action-value function $Q(s,a)$, which learns the expected reward from taking action $a$ in state $s$.
The action-value function generalizes over the set of all states so that it can handle previously unseen states.
The reinforcement learning agent can be subdivided into an actor, which learns the policy, and a critic, which learns the action-value function. 
The actor learns from the feedback of the critic and the critic learns from the feedback of the environment, the reward. 

Deep Deterministic Policy Gradient (DDPG) \cite{Lillicrap2016_DDPG} is an actor-critic RL algorithm for continuous actions and therefore useful for robotic control. 
It combines deep Q-networks \cite{Mnih2015} and deterministic policy gradient algorithms \cite{Silver2014}. 
DDPG trains the actor and critic neural networks off-policy-like, 
and stores the collected experience in a replay buffer. 

The actor-critic approach is often applied in a multi-goal framework, where the agent learns a policy that is conditioned to a goal variable \textit{g} \cite{Lillicrap2016_DDPG,Eppe2019_CGM,Roder2020_CHAC,Schaul2015_UVFA}. 
This allows for specifying different goals for the same task. For example, $g$ can specify the desired goal coordinates of its end-effector \cite{Eppe2019_planning_rl}, but it can also specify the desired note for a note-playing task.
Universal value function approximators (UVFAs) \cite{Schaul2015_UVFA} learn a value function $Q(s, a, g)$ that consider this goal variable, computing the expected value of taking action $a$ in state $s$ given the goal $g$.
Consequently, universal value functions do not only generalize over continuous states, but also over continuous goal representations.

A general problem of reinforcement learning is sample-efficiency. 
Hindsight experience replay (HER) \cite{Andrychowicz2017a} increases the sample efficiency for goal-conditioned RL by enabling the agent to learn from unsuccessful episodes. 
By substituting the goal in the replay buffer with the goal that has been actually achieved, the agent pretends that the achieved state was the desired goal. It then receives a reward for its actions and learns how to achieve the achieved state, pretending in hindsight that this was the goal.
This is especially helpful in situations where the rewards are sparse, and where it is unlikely for the agent to receive a reward during exploration. In this work, we extend HER to temporally extended goals. 

\begin{figure*}[!b]
    \centering
    \includegraphics[width=\linewidth]{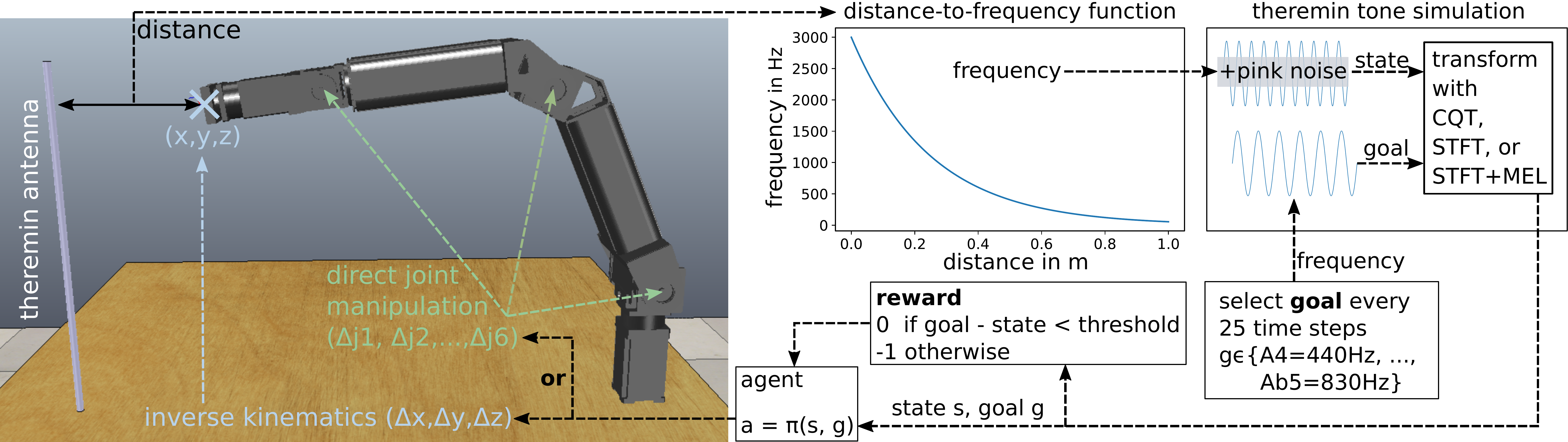}
    \caption{Schematic depiction of the agent-environment interaction loop. The simulated 6 DOF robotic arm and the theremin antenna are illustrated on the left. State and goal are transformed to the frequency domain either by the CQT, the STFT or the STFT with mel scaling. The robot is actuated with inverse kinematics (blue) or direct manipulation of the joints (green).}
    \label{fig:env_scheme}
\end{figure*}

Reinforcement learning has been used for the synthesis of music-like signals by Doya and Sejnowski \cite{Doya1995}. The authors modeled song-learning by birds with a reinforcement learning model that chooses the para\-meters for a synthesizer to replicate recordings of birdsongs. 
Their work has many similarities to our approach to operate musical robots: 
they train their model only with auditory feedback and evaluate the agent's actions one syllable at a time.
In our work, we extend these ideas and, rather than using a synthesizer, let the agent interact physically with the environment. Moreover, we extend UVFAs towards temporally extended goal variables so that the agent also learns to generalize over environment states and temporally extended goals.


\section{Methodology}
\label{sec:meth}

\subsection{Time-dependent Goals}

Classical UVFAs enable an agent to learn to play a single note on an instrument. However, such an agent is not designed to learn to play sequences of notes because it only considers one goal per episode.
Our proposed time-dependent goal enhances UVFAs by defining a possibly different goal $g_t$ for every time step $t$. 
This is necessary to describe the task of playing music because melodies are a sequence of notes that have to be played at the right time.
To integrate our proposed time-dependent goal (TDG) with actor-critic RL, we make several adjustments.
We start by changing the signature of the reward function from $r(s_t, g)$ to $r(s_t, g_t)$. This results in the following modified Bellman equation (\ref{eq:bellman}) for the optimal action-value function:

\begin{equation}
    \label{eq:bellman}
    \begin{split}
        Q^*(s_t,a_t,g_t) & = r(s_{t+1}, g_{t+1}) \\&+ \gamma \cdot \underset{a}{\mathrm{max}} \; Q^*(s_{t+1}, a_{t+1}, g_{t+1})
    \end{split}
\end{equation}

Herein, $Q^*(s_t, a_t, g_t)$ is the optimal action-value function and $\gamma$ is the discount factor for future rewards.
In our work, we implement the actor as an artificial neural network $\pi$ with weights $\theta$, and we implement the critic $Q$ one with weights $\phi$. Hence, we rewrite the reinforcement learning problem with TDGs as the following optimization problem (\ref{eq:crit_opti}) for the critic's weights:

\begin{equation}
    \label{eq:crit_opti}
    \begin{split}
        \arg\min_\phi [r(s_t, g_t) &+ \gamma \cdot Q_{\phi}^{\pi_\theta}(s_t, a_t, g_t) \\ &- Q_{\phi}^{\pi_\theta}(s_{t-1}, a_{t-1}, g_{t-1})]
    \end{split}
\end{equation}

Accordingly, we optimize the actor's weights with the output of the critic as follows (\ref{eq:act_opti}):
\begin{equation}
    \label{eq:act_opti}
    \arg\min_\theta [Q_\phi^{\pi_\theta}(s_t, \pi_\theta(s_t, g_t), g_t)]
\end{equation}

Since we use an off-policy approach, we also store the TDG in the replay buffer. Therefore, the stored transitions are tuples ($s_t, a_t, g_t, s_{t+1}, g_{t+1}$).
In our experiments we show that these enhancements to goal-conditioned RL enable agents to learn to achieve sequences of time-dependent goals that encode a desired sequence of simulated theremin tones.

\subsection{Theremin Tone Simulation}
In our simulation, we approximate the real-world theremin properties and represent its sound in the frequency domain.
The pitch of the theremin is calculated from the distance of the arm's tip to the theremin antenna with a formula that approximates a real-world theremin distance-to-pitch relation (see Fig.\ref{fig:env_scheme}).
With the pitch frequency, we compute a 50 millisecond-long theremin tone $x(n)$ with base frequency $f$ that mimics a real-world theremin sound described by the following equation (\ref{toneeq}).
\begin{equation}
\label{toneeq}
x(n) = \sum_{i=0}^{8} 0.4 (1 / 4^i) \mathrm{sin}(2 \pi f (i+1)n)
\end{equation}
The theremin sound is composed of eight sine waves, such that their frequencies are multiples of the base frequency (harmonics).
The amplitude of each sine is scaled by 0.4 and a factor that exponentially decreases the amplitude of each harmonic the higher its frequency is.
We determined these parameters by manual numerical optimization. We only compute the first seven harmonics because higher harmonics have a relatively low amplitude. Therefore, the effect of higher harmonics is negligible.

We add pink noise with a signal-to-noise ratio (SNR) of 38~dB to the simulated theremin signal to improve the robustness of the agent.
We transform the tone into the frequency domain and compare the following three methods: Constant-Q transform (CQT) \cite{brown_cqt}, short-time discrete Fourier transform (STFT), and STFT with mel scaling. 
We use a hann-window and discard the phase for all transformations and provide the transformed signals as perceptive input to the agent.

\section{Experiments}
\label{sec:expe}

We utilize the time-dependent goal for training a thereminist robot in a simulation and transfer a policy to the real world. 
In both the simulation and real-world scenario the agent's task is to play a sequence of eight notes with the right duration on the theremin.

\subsection{Simulation}
We simulate a 6 DOF robotic arm and a theremin antenna in CoppeliaSim \cite{CoppeliaSim} and use PyRep \cite{James2019} to actuate the robotic arm.
Fig. \ref{fig:env_scheme} illustrates the agent-environment interaction loop. 
At each time step, two theremin tones are computed and provided to the agent in their frequency domain representation. One is the observation, whose frequency is determined by the distance between the robot's tip and the theremin antenna. The other one is the time-dependent goal.

One episode consists of 200 time steps and the time-dependent goal changes every 25 time steps. 
The goal-notes are randomly selected between A4 and Ab5.
The agent tries to achieve these goals by actuating the robotic arm with either inverse kinematics or by direct manipulation of the joints.
The sparse reward is computed with spectrographic template matching described in the following equation (\ref{eq:reward_formula}), where $g$ is the goal, $s$ is the achieved state and n\_bins is the number of bins of the used transform (constant-Q transform / short-time discrete Fourier transform (STFT) / mel-scaled STFT).
\begin{equation}
    \label{eq:reward_formula}
r(s, g) = \begin{cases}
0 & \sum_{i=0}^\mathrm{n\_bins}{\lvert g_{i} - s_{i} \rvert < \epsilon}\\
-1 & \text{otherwise}
\end{cases}
\end{equation}
The difference between the goal $g$ and the achieved state $s$ at time t has to be smaller than a threshold $\epsilon$ for the reward to be 0, otherwise the reward is -1.
We choose $\epsilon$ so that the relative difference between the base frequency of the goal note and the achieved goal note is smaller than 0.7\% of the goal note's frequency, which is larger than the difference trained humans can notice but smaller than the difference untrained humans notice \cite{Micheyl2006}.
At the start of each episode, the robot's position is reset and the position of the theremin antenna is randomized in the y-direction, which means that it is moved sideways.


\subsection{Real World}
\begin{figure}[t]
         \centering
          \includegraphics[width=\linewidth, trim=0 0 0 -0.8cm]{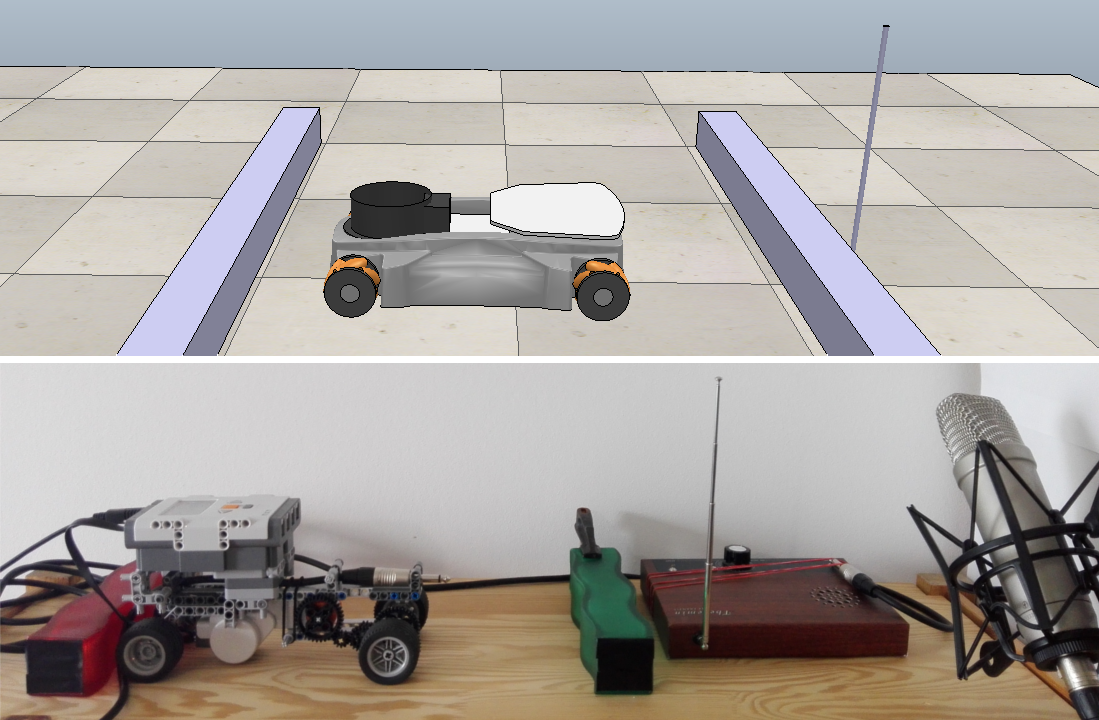}
         \caption{Simulation (top) and real-world (bottom) environment for the theremin playing robot. The robot's movement is constrained by two blocks. We also trained a 6 DOF robotic arm (See Fig.\ref{fig:env_scheme}) but did not have access to its physical version due to the COVID-19 pandemic.}
          \label{fig:real_env}
\end{figure}

The real-world robot is a 1 DOF mobile robot that can move back and forth, thereby changing its distance to the theremin and therefore the pitch.
We trained an agent in the simulation and then deployed it on the Lego Mindstorms NXT robot shown in Fig. \ref{fig:real_env}.

The real-world sound is recorded by a condenser microphone and transformed to the frequency domain with the CQT. The action is a value that specifies the desired rotation of the robot's motor.
To account for the fluctuating amplitude of the real-world theremin and the limited accuracy of the NXT robot, we change the reward computation during the simulated training as described in the following equation (\ref{eq:new_rew_func}). 
\begin{equation}
r(s, g) = \begin{cases}
0 & \mathrm{argmax} \; \mathrm{CQT}(g) == \mathrm{argmax} \; \mathrm{CQT}(s)\\
-1 & \text{otherwise}
\end{cases}
\label{eq:new_rew_func}
\end{equation}
The reward is now 0 if the CQT bin with the largest value corresponds to the goal note and -1 otherwise. This effectively increases the allowed tolerance when computing the reward. 
In contrast to the 6 DOF robot simulation, the position of the robot is not reset at the start of every episode. 
Instead, it just starts from the position it was at at the end of the last episode. This automatically makes the agent generalize over different starting positions.

\subsection{Agent Neural Network Architecture}

\begin{figure}[h]
         \centering
          \includegraphics[trim=0 0 0 -0.8cm, width=\linewidth]{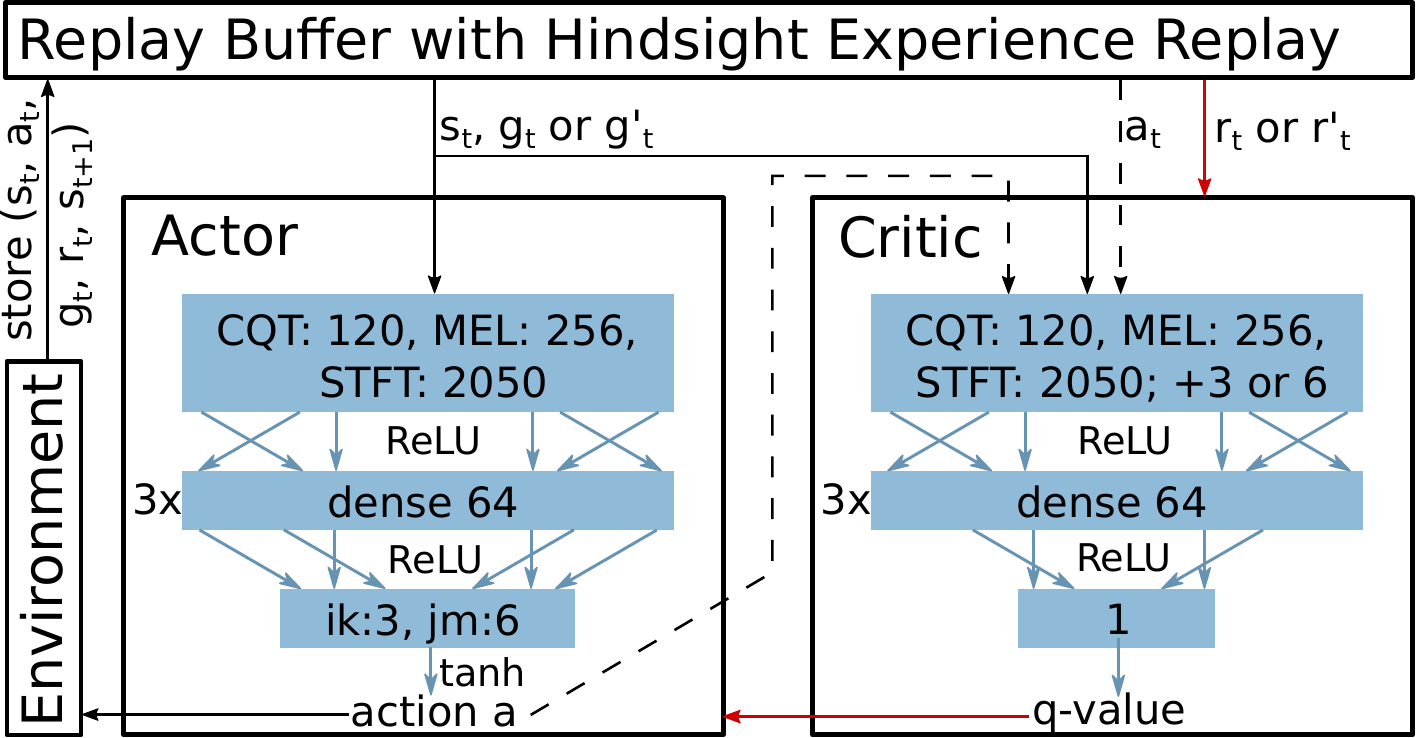}
         \caption{Agent architecture. Blue boxes denote neural network layers. Sizes of input and output layers depend on the transform type (CQT / mel-scaled STFT / STFT) and actuation type (inverse kinematics / joint manipulation). The left dotted line is used for the actor-update and the right dotted line is used for the critic-update. $g'_t$ is the goal chosen by HER and $r'_t$ is the recomputed reward for the substituted goal. The red arrows indicate the values that we used as feedback for the networks.
         }
          \label{fig:arch}
\end{figure}

The agent architecture is a DDPG actor-critic network.
Actor and critic both have three dense layers with 64 neurons and ReLU activation each, as illustrated in Fig.\ref{fig:arch}. 
We further increase the sample efficiency with hindsight experience replay \cite{Andrychowicz2017a} with a manipulated to original experience ratio of 4:1 and replay strategy \textit{future}, which means that HER-goals are replaced by an achieved state from a future time step. 
During training, we randomize 30\% of the actions and add 20\% action noise to the non-random actions to facilitate exploration. 
We use the Adam Optimizer \cite{Kingma2015_Adam} and a learning rate of 0.001.
The discount factor $\gamma$ for future rewards is 0.995 and the loss for the critic is the mean-squared Bellman error.
One episode consists of 200 steps and the time-dependent goal changes every 25 steps.
Our implementation builds upon the OpenAI baselines implementation of DDPG with hindsight experience replay \cite{baselines}.

\section{Results and Discussion}
\label{sec:resu}

Both the simulated and the real-world robot can successfully play the theremin. 
We evaluate the accuracy of the simulated robot by caching the base frequencies of the goals and achieved states and use them to calculate the amount of time steps at which the goal has been reached, so that the transform type does not influence the evaluation.

Our baseline configuration uses the CQT, inverse kinematics, a signal-to-noise ratio of 38~dB, and HER. In our experiments, we alter these to find out the best configuration.
We train each configuration seven times for 30 epochs, which consist of 25 train- and 10 test-episodes each. The graphs show the median and the second and third quartile of the training runs.
\begin{figure}[b]
         \centering
          \includegraphics[width=\linewidth]{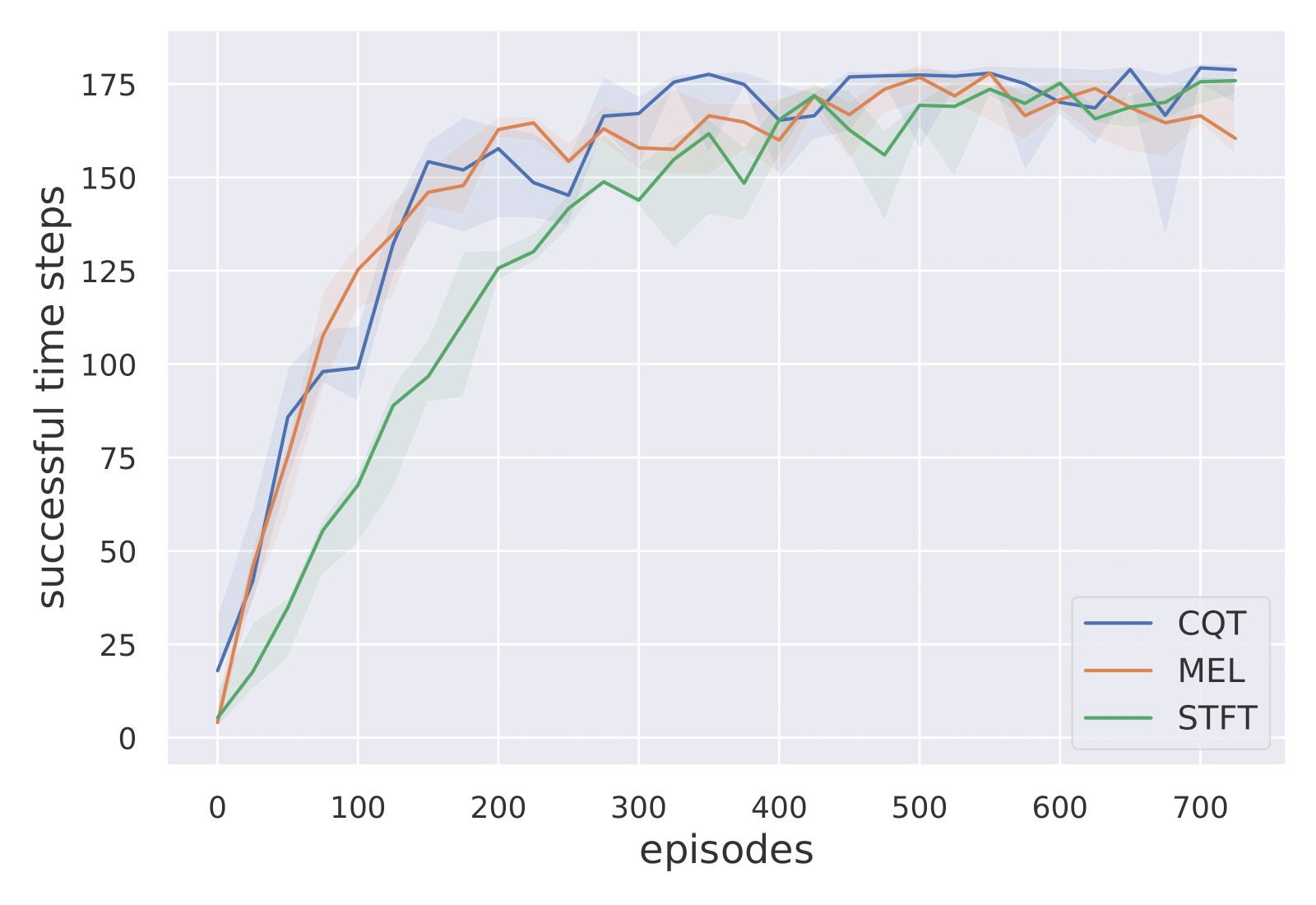}
         \caption{Performance of the agent using the different time- to frequency-domain transforms constant-Q transform, short-time discrete Fourier transform, and STFT with mel scaling. }
          \label{fig:sound_repr}
\end{figure}
\subsection{Transforms}
We compare the three different time- to frequency-domain transforms CQT, STFT, and STFT with mel scaling.
\autoref{fig:sound_repr} depicts the results.
The learning converges after approximately 300 - 500 training episodes, depending on the
transformation method used. 
CQT performs slightly better than the other methods because the spacing of the center frequencies is better suited for musical notes. 
The spacing is logarithmically arranged with a constant factor of $Q = \frac{1}{12}$ and, therefore, corresponds to the typical chromatographic note frequencies of Western music that we used. 


For all transformation methods, it is not possible to achieve 100\% accuracy. 
The successful time steps converge to around 175 after 300-600 episodes and never reach the full 200 successful time steps per episode because the agent has to move between the notes. While still in the process of moving, it has not yet hit the correct pitch. 

\subsection{Action Spaces}
\begin{figure}[b]
         \centering
          \includegraphics[width=\linewidth]{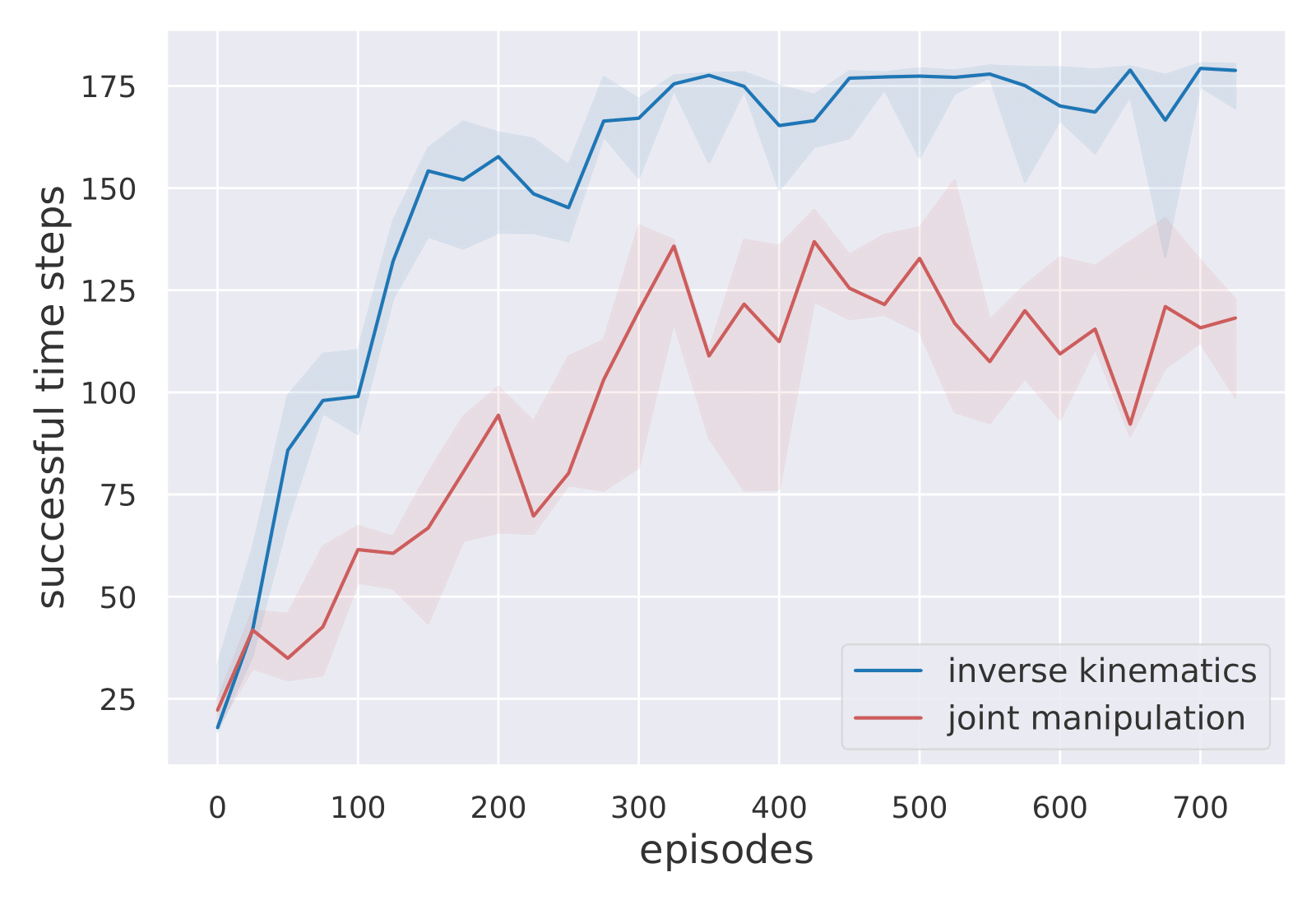}
         \caption{Performance of the agent using either inverse kinematics or direct joint manipulation for actuating the robot.\newline}
          \label{fig:JM}
\end{figure}

We also compare two action spaces of the robot's end-effector: First, we control the robot in Cartesian space with inverse kinematics, and, second, we perform direct control over the joints. Fig. \ref{fig:JM} shows that the agent performs better when it uses inverse kinematics.
This was expected because the Cartesian space is smaller, it directly correlates with the theremin's pitch, and the joint movements are not independent of one another.

\subsection{Need for Hindsight Experience Replay}
\begin{figure}[t]
         \centering
          \includegraphics[width=\linewidth]{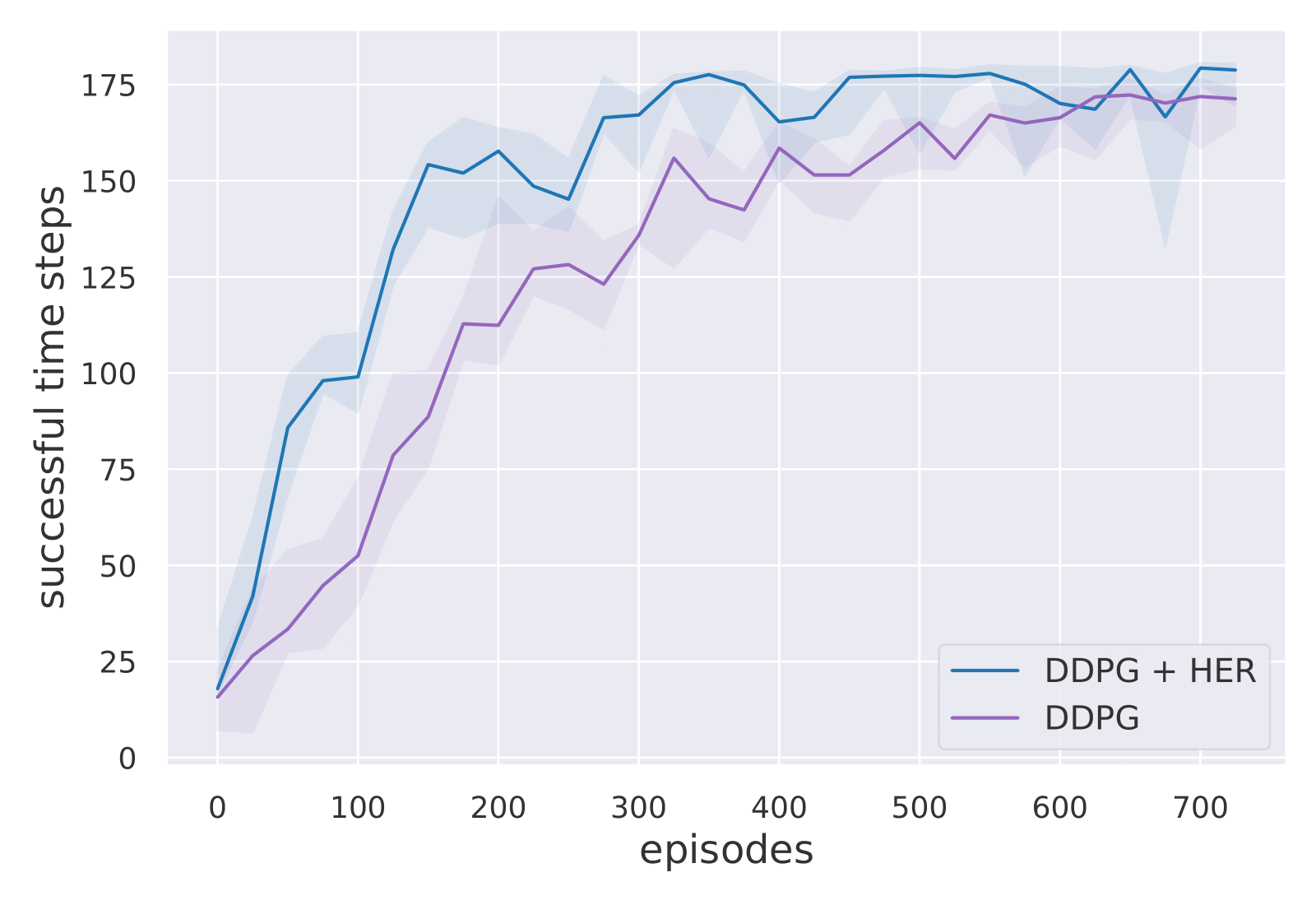}
         \caption{Influence of HER on the performance of the agent. \newline}
          \label{fig:noHER}
\end{figure}

We also scrutinize the benefits of using hindsight experience replay.
Our results (Fig.\ref{fig:noHER}) show that HER improves the training speed. However, the benefits of HER compared to DDPG alone are not as dramatic as expected. We hypothesize that the agent is likely to hit the right notes by accident, generating sufficiently many successful training samples for the training, so that DDPG without HER already performs well. 

\subsection{Robustness and Generalization}
\begin{figure}[t]
         \centering
          \includegraphics[width=\linewidth]{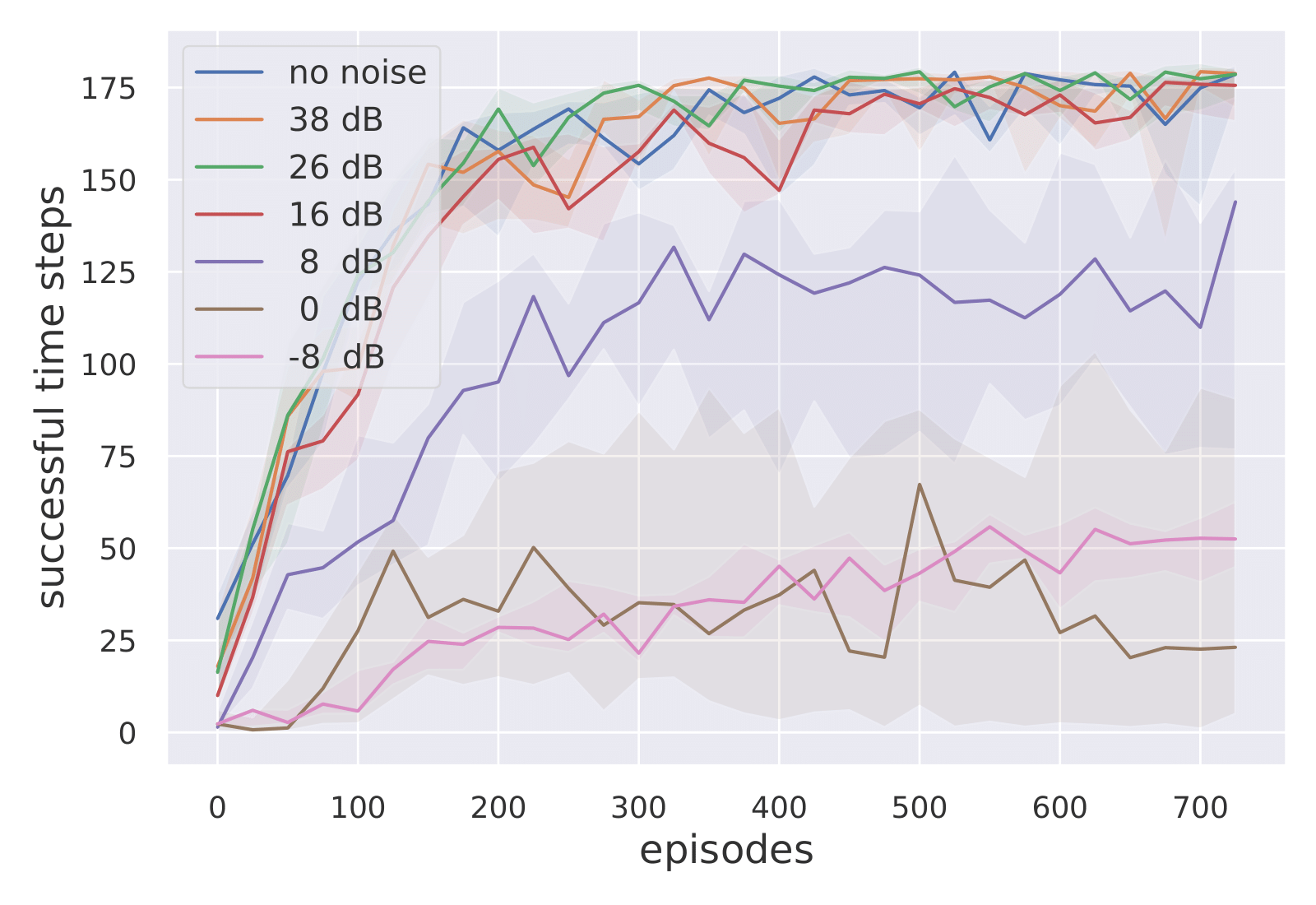}
         \caption{Performance of the agent at different signal-to-noise ratios.}
          \label{fig:SNR}
\end{figure}

Fig. \ref{fig:SNR} shows that different signal-to-noise ratios (SNRs) affect the performance of the agent, but the agent is rather robust to auditory noise. The agent performs well at SNRs up to 16~dB and still learns something when the noise is as loud as the signal. A large drop in performance only appears at a SNRs below 8~dB. 

In addition to this empirical evaluation, we perform two qualitative tests to further investigate the robustness of our approach. 
First, we evaluate the agent with tones that it was not trained on, i.e. with tones outside of the discrete chromatographic musical scale. In this case, the agent was still able to hit the correct frequencies. Second, we test a changed distance-to-pitch computation. This also does not hinder the agent from achieving the goals. 
The successful deployment of a policy to the real world shows the robustness of the agent as well.

\subsection{Real World}
To test the transfer to real-world robots, we train the NXT robot in simulation and deploy the learned policy on the physical robot.
The NXT is physically not able to play realistic melodies because it lacks speed and accuracy to achieve an appropriate timing performance.  
However, as illustrated in our supplemental video\footnote{\url{https://youtu.be/jvC9mPzdQN4}}
, the NXT robot can play a slow series of notes on a real theremin. This is possible even though the NXT's motor noise is relatively loud.
Furthermore, we observed that the robot can also react to and compensate for external disturbances, e.g. when we increase the theremin's pitch by moving a hand closer to the antenna. In this case, the agent drives backward to compensate for the disturbance.

\section{Conclusions}
\label{sec:conc}
We introduced the concept of time-dependent goals that enable robots to play musical instruments. We provided a proof of concept for the case of actor-critic reinforcement learning, but the method is equally applicable to other goal-conditioned reinforcement learning methods that are based on UVFAs. We demonstrated that a robotic agent can successfully learn to play the theremin in simulation and that the learned behavior can also be transferred to a physical robot. Our future work will involve more sophisticated physical robots and more complex instruments, such as a xylophone or a keyboard. 
In addition, we will extend the time-dependent goal method on a theoretical level, so that it can also account for future goals. This will enable the robotic agent to prepare for playing a note beforehand so that it can start the movement required to hit the note in time. 

\addtolength{\textheight}{-8.1cm}   


\section*{Acknowledgment}
Manfred Eppe, Stefan Wermter and Phuong Nguyen were supported by the German  Research  Foundation  (DFG)  through the IDEAS and LeCAREbot projects. The authors gratefully acknowledge partial support from the German Research Foundation (DFG) under project Crossmodal Learning (TRR-169). 

\small
\bibliographystyle{IEEEtranN}
\bibliography{referencesfinal}
\normalsize

\end{document}

%% file: preamble.tex
\IEEEoverridecommandlockouts                              

\usepackage[table]{xcolor}
\usepackage{graphicx} 
\usepackage{dblfloatfix}
\usepackage{amsmath} 
\usepackage{amssymb}  
\usepackage[disable]{todonotes}
\usepackage[hidelinks]{hyperref}
\usepackage{tikz}
\usepackage{caption}
\usepackage{subcaption}
\usepackage{multirow}
\usepackage{multicol}
\usepackage{makecell}
\usepackage{ctable}
\usepackage{xparse}
\makeatletter
\let\NAT@parse\undefined
\makeatother
\usepackage[numbers]{natbib}
\usepackage{fancyhdr}

\makeatletter
\def\NAT@spacechar{~}
\makeatother

\fancypagestyle{firstpage}{
  \lhead{Preprint, submitted to \textit{IEEE Robotics and Automation Letters (RA-L) 2021} with \textit{International Conference on Robotics and Automation Conference Option (ICRA)} 2021}
}